\documentclass[10pt, conference, compsocconf]{IEEEtran}
\ifCLASSINFOpdf
\else
\fi

\usepackage[utf8]{inputenc}
\usepackage[T1]{fontenc}
\usepackage{times}
\usepackage{epsfig}
\usepackage{graphicx}
\usepackage{amsmath}
\usepackage{amssymb}
\usepackage{url}

\usepackage{breakurl}
\usepackage[breaklinks]{hyperref}

\begin{document}
%
\title{Road User Detection in Videos}


\author{\IEEEauthorblockN{Hughes Perreault, Guillaume-Alexandre Bilodeau, Nicolas Saunier}
\IEEEauthorblockA{
Polytechnique Montréal\\
Montréal, Canada\\
Email: \{hughes.perreault, gabilodeau, nicolas.saunier\}@polymtl.ca}
\and
\IEEEauthorblockN{Pierre Gravel}
\IEEEauthorblockA{
Genetec\\
Montréal, Canada\\
Email: pgravel@genetec.com}
}


%


\maketitle

\begin{abstract}
Successive frames of a video are highly redundant, and the most popular object detection methods do not take advantage of this fact. Using multiple consecutive frames can improve detection of small objects or difficult examples and can improve speed and detection consistency in a video sequence, for instance by interpolating features between frames. In this work, a novel approach is introduced to perform online video object detection using two consecutive frames of video sequences involving road users. Two new models, RetinaNet-Double and RetinaNet-Flow, are proposed, based respectively on the concatenation of a target frame with a preceding frame, and the concatenation of the optical flow with the target frame. The models are trained and evaluated on three public datasets. Experiments show that using a preceding frame improves performance over single frame detectors, but using explicit optical flow usually does not.
\end{abstract}

\begin{IEEEkeywords}
object detection, video object detection, road users
\end{IEEEkeywords}

%
\IEEEpeerreviewmaketitle

\section{Introduction}
\begin{figure}[t]
\begin{center}
\includegraphics[width=0.8\linewidth]{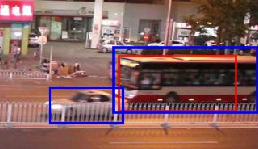}
\end{center}
   \caption{In this difficult case of occlusion, our model RetinaNet-Double detects the vehicle (blue) while the baseline does not. It is reasonable to think that the motion of the vehicle helped the model. }
\label{fig:diff}
\end{figure}

Automatic road user detection is used by an increasing number of applications. In the context of traffic monitoring and intelligent transportation systems (ITS) for example, detecting road users can provide traffic counts, speeds and travel times for traffic state estimation, and incident detection. The research presented in this paper is based on the following observations. First, it is reasonable to expect that most objects of interest are moving. Second, traffic monitoring images contain a large number of small vehicles, for example in the farthest areas of the camera field of view. Finally, given the often large number of vehicles on the road and urban furniture and typical camera positions, occlusion is a frequent phenomenon (an example of occlusion is shown in figure \ref{fig:diff}). All these challenges for road user detection could be better addressed by considering multiple frames instead of a single frame at a time. 

Current research in object detection is mostly focused on single frame detectors, even though many applications provide video streams. The most widely known state-of-the-art detectors work with a single frame at a time, while detection performance could be improved with some modifications.  

In this work, a novel approach for road user detection in videos is developed and evaluated. The proposed approach is generic and can be integrated into most existing object detection methods. We show that by using two frames, one does not need to compromise on speed, and detection accuracy can be improved. 

In the first model, we train the network to learn how to combine two frames of a video to perform object detection. We do that by concatenating the two images and feeding the result as an input to the deep network that performs the feature extraction. In our second model, we feed the network with the concatenation of the target image and its optical flow. The second model tries to accelerate the learning process, and in a way assumes that what the first proposed model learns is to associate detected movement with the presence of an object. Results show this might not be all there is to it.

The two models are trained, evaluated and compared with a baseline in different training settings. Our contribution is demonstrated through the improved performance of the proposed models over the baseline on three road user datasets. 

This paper is organized as follows: first we will present a brief literature review in section \ref{rwork}, then our models and contributions will be explained in section \ref{pmethod}, we will follow by presenting the datasets we used and the experiments we conducted in section \ref{exper}, before discussing our results in section \ref{resul} and finally concluding in section~\ref{conc}.  

\section{Related Work and Background}
\label{rwork}
Object detection in a single image has been widely studied and is the subject of many research papers every year. There has been some major breakthrough on this topic in the recent years. The classical methods, those created before the deep learning era, got surpassed by a large margin. This review will focus on deep learning-based methods. For single frame object detection, there are two main categories: the one-stage approach and the two-stage approach. The difference between the two is that the two-stage approach uses an object proposal phase in which they select the best candidates for further processing, and the one-stage approach detects objects directly. We will then review object detection in videos. We will finish the literature review by presenting some work that has inspired us for this work, on optical flow estimation by deep neural networks (DNNs). 

\textbf{Convolutional neural networks: }convolutional neural networks (CNNs) have caused a revolution in the field of object detection. AlexNet~\cite{krizhevsky2012imagenet} was the first network to beat classical methods on ImageNet~\cite{imagenet_cvpr09}, by using clever strategies such as dropout~\cite{srivastava2014dropout}, batch normalization~\cite{ioffe2015batch} and rectified linear units. VGG~\cite{Simonyan2014} was also extremely influential, as it established good practices by its simplicity and elegance. For deeper networks, ResNets~\cite{he2016deep} use skip connections to build a network out of residual blocks, which help propagate the gradient. This allows ResNets to go as deep as 150~layers. For faster networks, MobileNets~\cite{howard2017mobilenets} use the idea of depth separable convolutions in order to speed up the computation process and save memory. The result is a network suitable for vision applications that can run on mobile devices. 

\textbf{Two-stage approach: }two-stage detectors have a separate object proposal phase. They were the first ones to incorporate DNNs in their architecture. R-CNN~\cite{Girshick_2014_CVPR} used selective search~\cite{uijlings2013selective} to get object proposals, and used a CNN, such as VGG, to classify each proposal. This architecture was accurate but very slow due to having to run the whole DNN on every single proposal. R-CNN was very influential and was the first of a family of state-of-the-art object detectors. Several improvements were then proposed to make these two stages completely trainable end-to-end and to make them share most of the computation. Fast R-CNN~\cite{Girshick_2015_ICCV} solves the bottleneck of having to run each proposal through the entire DNN by introducing an ROI pooling layer that extracts the relevant features for each proposal out of the feature map of the whole image. That way, the DNN only has to run once over the whole image. After that, Faster R-CNN~\cite{NIPS2015_5638} solved the limitation of having to use an external object detector by introducing the RPN, a DNN that generates object proposals. The RPN shares the vast majority of its layers with Fast R-CNN, creating a unified and end-to-end trainable object detector. Several variants of these architectures were released, each to work on one or multiple specific problems, for example multi-scale detection~\cite{Lin_2017_CVPR}, and a faster two-stage detector~\cite{NIPS2016_6465}.

\textbf{One-stage approach: }the one-stage approach addresses the bottleneck of the two-stage approach, the computation required for each proposal. It does that by simply removing the object proposal phase, thus the name. The first accurate one-stage object detector was also the first to work at a real-time speed, YOLO~\cite{Redmon_2016_CVPR,Redmon_2017_CVPR}. YOLO works by dividing the image into a regular grid, and having each cell of the grid predict two bounding boxes for objects. The loss function of YOLO is a combination of a classification and localization loss. SSD~\cite{liu2016ssd} proposed an improvement on YOLO by using a simpler architecture. Notably, SSD introduced the idea of using anchor boxes on the feature maps, as in the region proposal network of Faster R-CNN~\cite{NIPS2015_5638}. In SSD, a sliding window is performed on the feature maps using the anchor boxes, and detection is done at every single location. RetinaNet~\cite{lin2018focal} works in a very similar fashion as SSD, but introduces a new loss function dubbed ``focal loss'', that aims to correct the imbalance in background and foreground examples of one-stage detectors during training. To improve multi-scale detection, RetinaNet also uses a pyramid of features and detects at multiple levels of this pyramid. To this day, it is impossible to state whether one family of object detectors has won over the other, and the competition for the best and faster detector is ongoing. 

\textbf{Object detection in video: }contrarily to object detection on single images, this task has seen less research. Here is presented some of the most notable recent works on this topic. Recently, Liu \& Zhu~\cite{Liu_2018_CVPR} used a Long short-term memory (LSTM) to propagate and refine feature maps between frames, allowing them to detect objects a lot faster while keeping a precision similar to a single frame detector. Before them, Zhu et al.~\cite{Zhu_2017_CVPR} used the optical flow information to propagate feature maps to certain frames in order to save computation time. These two articles focus on reusing computation between frames in order to save time based on the premise that there is a lot of similarity and continuity between frames. In comparison, our work focuses on improving the precision and recall of the detection by combining information. By using deformable convolutions instead of optical flow training, Kim et al.~\cite{Bertasius_2018_ECCV} trained a model to compute an offset between frames, and thus are able to sample features from close preceding and following frames to help detect objects in the current frame. Their model is particularly good in cases of occlusion or blurriness in the video.  

\textbf{Optical flow by DNNs: }one of the most notorious work in optical flow learning is without a doubt FlowNet~\cite{Dosovitskiy_2015_ICCV}. This work presented the first end-to-end network that learns to generate optical flow from a pair of images. They proposed two models, FlowNetSimple and FlowNetCorr, that are both trained on a digitally constructed dataset with 3D models of chairs. The dataset was created by moving these chairs on different backgrounds. The models take as input a pair of consecutive images. FlowNetSimple works by simply concatenating images and letting the network learn how to combine them. This is the inspiration for one of our model. FlowNetCorr, on the other hand, works by computing a correlation map between the high level representation of the two images. A second version, FlowNet~2.0~\cite{Ilg_2017_CVPR} proposed several improvements over FlowNet, most notably to stack several slightly different architectures one after the other to refine the flow. 

\section{Proposed Method}
\label{pmethod}

\subsection{Problem Statement}
The task that we want to solve is as follows: given a target image, a preceding frame and a set of labels, locate with a bounding box and classify every object in the target image that corresponds to one of the labels. Using the immediately preceding frame is not mandatory, but the model may not use any future frames in order to work online. 

\subsection{Overview}
\begin{figure*}[t]
\begin{center}
\includegraphics[width=0.8\linewidth]{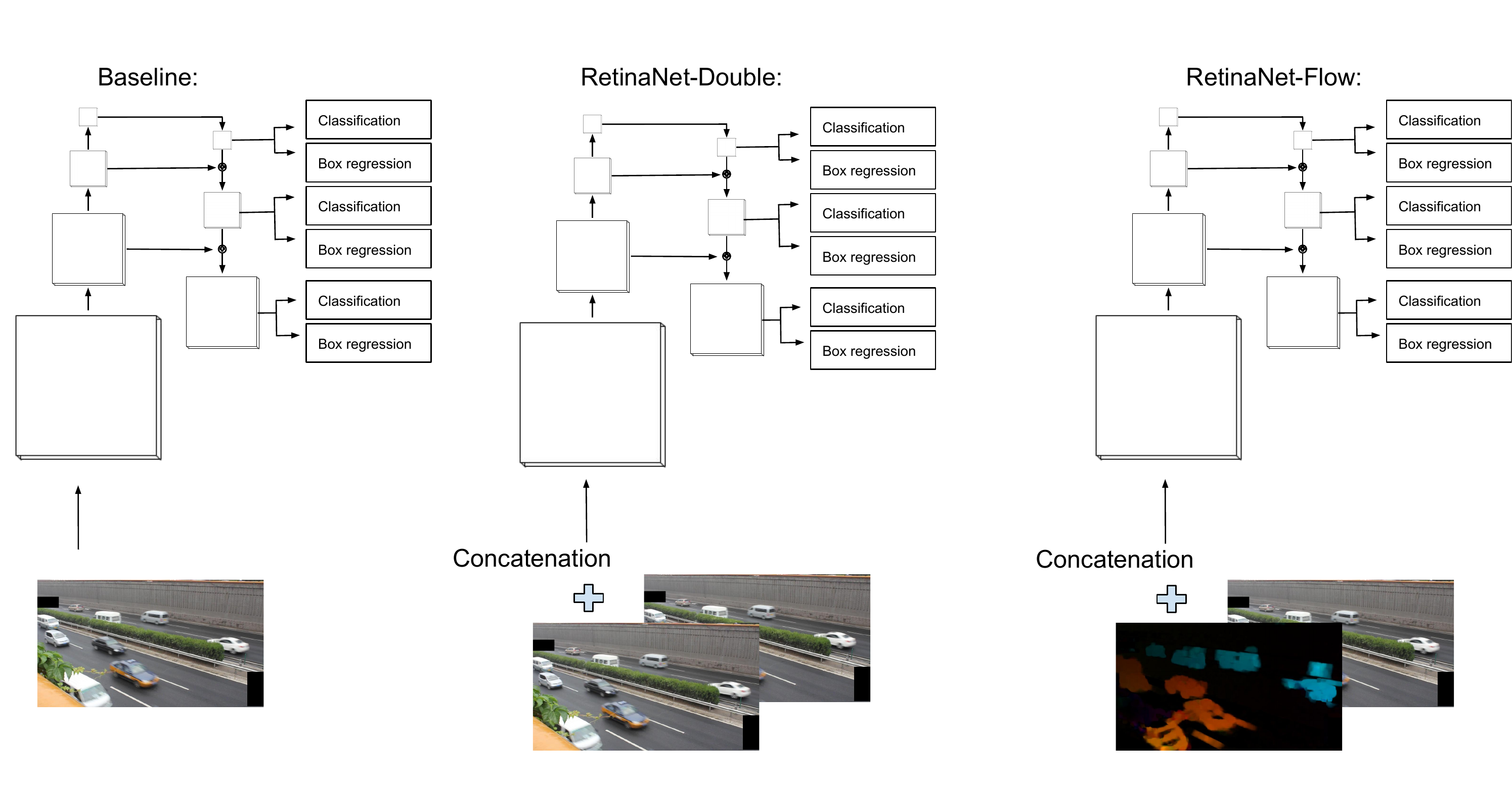}
\end{center}
   \caption{Presentation of our compared models: the first one is our baseline, a simple RetinaNet, the second one is RetinaNet-Double, which uses as input a concatenation of consecutive images, and the third one is RetinaNet-Flow, which uses as input a concatenation of the optical flow and the target image. RetinaNet builds feature maps at different resolutions and use a sliding window approach on these feature maps with different anchor boxes to find objects. These anchor boxes pass through a classification and a regression sub network.}
\label{fig:model}
\end{figure*}

To capitalize on consecutive frames in the video setting, the input stream of RetinaNet was changed using concatenation (see figure~\ref{fig:model}). The added inputs are either optical flow or a preceding frame. When using a preceding frame, the idea is that the network will learn by itself the best way to combine the two images and extract features for object detection. The motivation to use a direct concatenation and not two streams concatenated later comes from FlowNetSimple~\cite{Dosovitskiy_2015_ICCV}, where it was shown to be sufficient to train a CNN to learn motion. When using optical flow, the best information the network can learn is assumed to be movement, which is fed directly to it. By using two different models as well as the baseline, we have a good way of comparing them and finding out the best for the different kind of classes or cases.

\subsection{Baseline: RetinaNet}
In this work, multiple architectures were implemented and tested. The baseline is the state-of-the-art architecture RetinaNet~\cite{lin2018focal} with VGG-16~\cite{Simonyan2014} and ResNet50~\cite{He_2016_CVPR} as backbone feature extractors. We used both alternatively in our experiences in order to show that our models were not dependant on a specific feature extractor. RetinaNet is a model that uses a CNN to create a feature pyramid network, that is, a pyramid of feature maps at different scales. On each of the pyramid levels, it runs a sliding window with multiple anchor boxes at different scales and aspect ratio, and it runs these boxes into a classification and a box regression sub-network. It will keep the best of these detections as its final output. 
RetinaNet was chosen over other architectures due to its relatively high speed and very good performance. For training the model, the adam optimizer from Keras~\cite{chollet2015keras} was used, with a smooth L1 loss for regression, and the focal loss
\begin{equation}
    FL(p')=-\alpha_t(1-p')^\gamma log(p')
\end{equation}
 for classification, where $\gamma$ can be seen as a factor that reduces the contribution of easy examples to the loss. In this work, $\gamma$ is set to 2. $\alpha_t$ is the inverse class frequency, and is used so that underrepresented classes have more weight in the training. $p'$ is the probability $p$ of predicted label if it corresponds to the ground-truth label, and is $1-p$ otherwise. Intuitively, this means that if the predicted probability is high and the prediction is correct, or if the predicted probability is low and the prediction is incorrect, the loss will be mainly unaffected. These are the easy examples. Otherwise, the example is considered hard and the loss will be amplified. The initial learning rate was set to 1e-5. 

\subsection{RetinaNet-Double}
The first proposed architecture is dubbed RetinaNet-Double. Inspired by FlowNetSimple~\cite{Dosovitskiy_2015_ICCV}, the architecture of the backbone model was modified so that it take as input two images that have been concatenated channel-wise, resulting in a DNN that takes a six-channel image as input. The rest of the RetinaNet model is left unchanged. This results in a model that is only slightly slower than its counterpart, as the number of parameters is practically the same. 

Our goal is to make the network learn how to combine data from two images to improve its performance. It is trained end-to-end on detection accuracy, and thus learns the features most useful for object detection and classification. It is not trained with optical flow ground-truths, but rather with detection ground-truths, meaning that the bounding box annotations and labels of the target image are used. This is the goal: the network must be trained with the standard classification and regression loss in order to learn how to correlate both images to perform better detection.

More formally, given a target image $I_t$ and a preceding frame $I_p$, the input of the network is constructed as 
\begin{equation}
    Input = Concatenate(I_p, I_t)
\end{equation}
which results in a tensor of shape:
\begin{equation}
    w\times h\times 2*c
    \label{eq:shape}
\end{equation} where $w$, $h$ and $c$ are respectively the width, height and number of channels of $I_t$ and $I_p$.

\subsection{RetinaNet-Flow}
The second proposed architecture, dubbed RetinaNet-Flow, uses external optical flow data generated with OpenCV~\cite{opencv_library}, specifically with the Farneback optical flow method~\cite{farneback2003two}. The model is similar to RetinaNet-Double, but takes as input the concatenation of the target image and a dense optical flow image. The network learns to associate movement with object presence. Dense optical flow is generated using the target image and a previous one, and the resolution of the resulting flow image is the same as the target. No optical flow ground-truth is used for training. We keep the X and Y components as well as the norm of the flow vector for each pixel. 

Again, given $I_t$ and $I_p$, we construct the input of RetinaNet-Flow by using
\begin{equation}
    Input = Concatenate(DenseFlow(I_p, I_t), I_t)
\end{equation}
where the shape of the resulting tensor is the same as equation~\ref{eq:shape}. 

\subsection{Choice of preceding frame}
We conducted an experiment to determine which preceding frame yielded the best results. This problem is defined as finding the best $i$ if the target frame is at time $t$, and the preceding frame at time $t - i$. We trained three models on UA-Detrac with $i=1$, $i=3$ and $i=5$. Our conclusion is that since the three models did not differ significantly on the validation set, it is better to use $i=1$ as it is the less restrictive and most intuitive of all the possible $i$'s. Also, in a real-time application, it would allow us to keep fewer images in memory. 

\section{Experiments and Results}
\label{exper}

\subsection{Datasets}
To train the proposed models, consecutive images from videos are needed. The models were trained and evaluated on three datasets: KITTI ``3 temporally preceding frames''~\cite{Geiger2012CVPR}, UA-Detrac~\cite{Wen2015Tracking} and the Unmanned Aerial Vehicle Benchmark (UAV)~\cite{du2018unmanned}. These were chosen specifically because they contain consecutive frames of moving road users.

\begin{figure}[t]
\begin{center}
\includegraphics[width=0.8\linewidth]{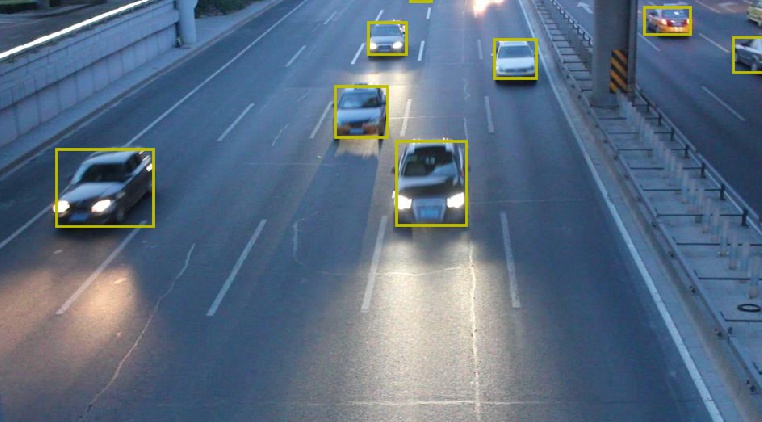}
\end{center}
   \caption{Example frame of UA-Detrac and its ground truth annotations.}
\label{fig:ua}
\end{figure}

UA-Detrac is a dataset created to benchmark object detection models and multi-object tracking in the context of traffic surveillance (Fig.~\ref{fig:ua}). All images have a 960x540 pixel resolution, and the models were trained on that resolution. Data is organized by video sequence with a fixed camera pointed on a road. The dataset contains 70000~images annotated with class labels and bounding boxes. There are four different labels in total: car, bus, van and other.  

\begin{figure}[t]
\begin{center}
\includegraphics[width=0.8\linewidth]{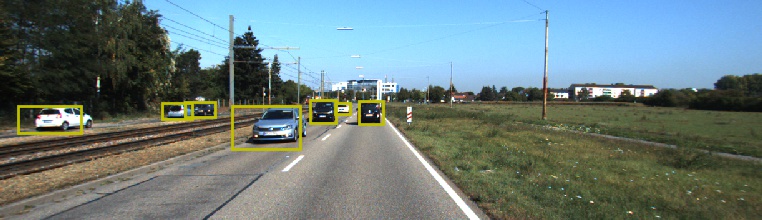}
\end{center}
   \caption{Example frame of KITTI and its ground truth annotations.}
\label{fig:kitti}
\end{figure}

KITTI is a dataset containing street-level images that can be used to train autonomous driving systems (Fig.~\ref{fig:kitti}). Images are 1224x370 pixels, to contain all the information useful for driving. The subset that we used, ``3 temporally preceding frames'', contains approximately 7500~triplets of consecutive frames. Eight classes are present in this dataset, Car, Cyclist, Misc, Pedestrian, Person\_sitting, Tram, Truck and Van.  

\begin{figure}[t]
\begin{center}
\includegraphics[width=0.8\linewidth]{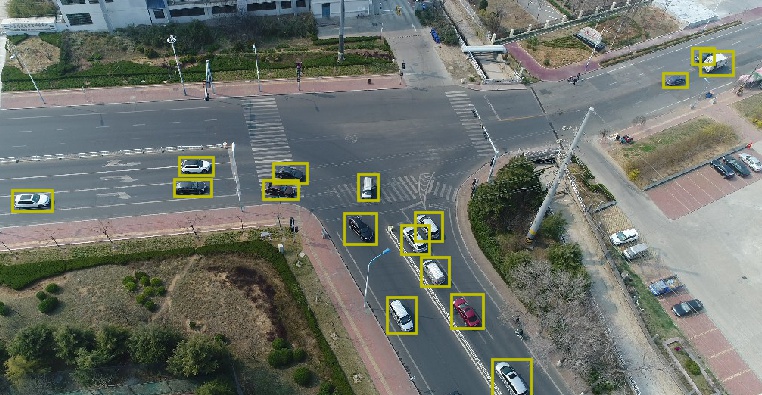}
\end{center}
   \caption{Example frame of UAV and its ground truth annotations.}
\label{fig:uav}
\end{figure}

UAV is a dataset of traffic scene videos in various conditions of weather, altitude and occlusion obtained by drones (Fig.~\ref{fig:uav}). It contains about 80000~annotated video frames. This dataset is particularly challenging due to its large number of small objects, high vehicle density and camera motion.  

UA-Detrac was randomly split into training and validation sets, with a ratio of 80~\%, 20~\% respectively. The models are evaluated remotely on a server, on their own test set. 
KITTI was randomly split in training, validation, and test sets, with a ratio of 60~\%, 20~\% and 20~\% respectively. The class distribution stays roughly the same in every set. Training is done using only the training set, and overfitting and accuracy are monitored on the validation set during training. Results are reported on the test set, for which models are only run once for evaluation. For UAV, we chose the same training / test split proposed in the development kit of the project, and we evaluated our results with this development kit as well.

\subsection{Experiments}
For a thorough evaluation, our proposed models were trained and evaluated in different ways. For the first training case, we trained every model from scratch, using randomly initialized weights and compare them on UA-Detrac and KITTI. In the second training case, we used transfer learning from UA-Detrac to KITTI. In that case, the last classification layer is skipped since the datasets do not contain the same number of classes. Finally, we train our models starting from pre-trained weights on ImageNet~\cite{imagenet_cvpr09} and compare them with state of the art models on the challenging UAV dataset. On that dataset, the three models were also trained from scratch in order to have a fair comparison between them. 

Indeed, for fairness, pre-trained weights should not be used for comparing with the baseline since the proposed models are new architectures for object detection that have to learn from scratch. The experiments were focused on comparing the proposed models with the RetinaNet baseline model trained in the same conditions as the new models. Since the datasets used are relatively small, the results are lower that what can be seen on benchmarks where pre-trained weights are used. RetinaNet-Flow was not trained on the KITTI dataset since the camera is moving for most examples and the resulting optical flow is not appropriate for object detection.  

\subsection{Results} 
All results reported in tables~\ref{tab:results1}, \ref{tab:results2} and~\ref{tab:results3} have been obtained with a minimum IOU of 0.7 for UA-Detrac and UAV, and 0.5 for KITTI as defined in their evaluation protocols. The IOU is the intersection over union, or the Jaccard index. The IOU between two rectangles is defined as the area of their intersection divided by the area of their union. The mAP computed for both datasets is the mean on all labels of the AP, the AP being the average precision given the recall and precision curves, therefore the area under the precision-recall curve. The backbone feature extractor used for KITTI and UAV is VGG-16~\cite{Simonyan2014} and ResNet50~\cite{He_2016_CVPR} is used for UA-Detrac. 

\begin{table*}[t]
\small
\setlength\tabcolsep{3pt} 
\def\arraystretch{1.5}
\centering
\caption{mAP reported on the UA-Detrac test set, for our two proposed models and the baseline. RN stands for RetinaNet, D for double and F for flow. Epochs is the number of epochs trained. }
\label{tab:results1}
\vspace{1em}
\begin{tabular}{c|c|c|c|c|c|c|c|c|c}
Model & Epochs & Overall & Easy & Medium & Hard & Cloudy & Night & Rainy & Sunny \\
\hline
\hline
RN-D-from-scratch & 20 &  \textbf{54.69}\% &	\textbf{80.98}\%	& \textbf{59.13}\%	& \textbf{39.23}\%	& \textbf{59.88}\%	& \textbf{54.62}\%	& \textbf{41.11}\%	& \textbf{77.53}\% \\
\hline
RN-from-scratch (baseline) & 20 & 46.28\% & 67.79\% &	49.42\% &	34.47\% & 	55.92\%	& 40.99\%	& 37.39\%	&56.43\% \\
\hline
RN-F-from-scratch & 20 & 40.70\% & 60.38\% & 44.94\% & 28.57\% & 48.94\% & 34.97\% & 32.43\% & 54.80\% \\
\end{tabular}
\end{table*}

\begin{table*}[t]
\small
\setlength\tabcolsep{3pt} 
\def\arraystretch{1.5}
\centering
\caption{mAP reported on the KITTI test set with (from-ua-detrac) and without (from-scratch) transfer learning. RN stands for RetinaNet and D for double. Epochs is the number of epochs trained. mAP is the mean of the AP over all classes. Under each class name is the AP for that class. }
\vspace{1em}
\label{tab:results2}
\begin{tabular}{c|c|c|c|c|c|c|c|c|c|c|c}
model & Epochs & mAP & Car & Cyclist & Misc & Pedestrian & Person\_sitting & Tram & Truck & Van \\
\hline
\hline
RN-D-from-scratch & 130 & \textbf{72.08\%} & \textit{86.95\%} & \textbf{56.80\%} & \textit{67.23\%}& \textbf{53.57\%} & \textbf{49.67\%} & \textit{89.21\%} & \textit{92.19\%} &\textit{ 81.00\%} \\
\hline
RN-from-scratch (baseline)& 130 & \textit{71.61\%} & \textbf{87.12\%} & \textit{54.67\%} & \textbf{69.67\%} & \textit{51.06\%} & \textit{45.13\%} & \textbf{90.04\%} & \textbf{93.35\%} & \textbf{81.82\%} \\
\hline
\hline
RN-D-from-ua-detrac & 60 & \textbf{68.29\%} & \textbf{87.59\%} & \textbf{51.15\% } & \textbf{63.68\% } & \textbf{54.30\%} & \textit{39.93\%} & \textit{85.62\%} & \textbf{90.42\%}  & \textit{73.63\%} \\
\hline
RN-from-ua-detrac (baseline) & 60 & \textit{67.68\% } & \textit{87.26\%} & \textit{46.67\%} & \textit{60.18\%} & \textit{47.16\%} & \textbf{44.93\% }& \textbf{87.93\%} & \textit{89.73\%} & \textbf{77.60\%} \\
\end{tabular}
\end{table*}

\begin{table}[t]
\small
\setlength\tabcolsep{3pt} 
\def\arraystretch{1.5}
\centering
\caption{mAP reported on the UAV test set. RN stands for RetinaNet, D for double and F for flow. The models are either trained from scratch when mentioned, or use pre-trained weights on ImageNet~\cite{imagenet_cvpr09} otherwise.}
\label{tab:results3}
\vspace{1em}
\begin{tabular}{c|c}
Model & Overall \\
\hline
\hline
R-FCN~\cite{NIPS2016_6465} & \textbf{34.35}\%\\
\hline
SSD~\cite{liu2016ssd} & 33.62\%\\
\hline
RN (baseline)& 33.48\%\\
\hline
RN-D & 31.15\%\\
\hline
RN-F & 30.41\%\\
\hline
Faster-RCNN~\cite{NIPS2015_5638} & 22.32\%\\
\hline
RON~\cite{kong2017ron} & 21.59\%\\
\hline
\hline
RN-D-from-scratch& \textbf{26.88}\%\\
\hline
RN-from-scratch (baseline)& 26.28\%\\
\hline
RN-F-from-scratch& 24.87\%\\
\end{tabular}
\end{table}


Table~\ref{tab:results1} presents the results obtained on the UA-Detrac dataset. One can notice that when trained from scratch, RetinaNet-Double shows a significant improvement over the baseline of 8~\%~points. However, RetinaNet-Flow did not perform as expected and scored 6\%~points lower than the baseline on the test set. We can interpret this result in a few ways. The network can make a good use of two frames in order to make better predictions. However, there might be too much noise in this dataset to use directly optical flow for detection, like trees moving in the wind and pedestrians on the side of the roads. An interesting finding is that it is better to let the network learn end-to-end how to combine frames instead of feeding it optical flow directly. 


Table~\ref{tab:results2} presents the results obtained on the KITTI dataset. One can see that when training from scratch, RetinaNet-Double achieves better results than the baseline, especially on the smaller objects like the classes with classes. For the Cyclist, Pedestrian and Person\_Sitting classes, the proposed model significantly outperforms the baseline model. When using pre-trained weights from UA-Detrac, the proposed model can achieve similar results by training approximately for about half the time of the training from scratch, with the same mAP difference in the results, which confirms the advantages of RetinaNet-Double. 

Table~\ref{tab:results3} presents the results obtained on the UAV dataset. While looking at the results for the models using pre-trained weights on ImageNet~\cite{imagenet_cvpr09} (RN (baseline), RN-D, RN-F), one must keep in mind that the very first layer of both RN-D and RN-F did not used pre-trained weights, putting them in a somewhat unfair comparison setting. Even so, we can see that our models are still competitive in terms of accuracy in this very challenging practical setting, outperforming Faster R-CNN but not surpassing the pre-trained SSD and R-FCN. The potential advantage of using pre-trained weights can be seen by comparing RN (baseline) and RN-from-scratch (baseline), and this also explains why RN (baseline) obtains better results than RN-D and RN-F. When looking at the results for the models trained from scratch, we can see that RetinaNet-Double outperforms RetinaNet again by 0.6\%~point mAP. However, RetinaNet-Flow still performs lower than the baseline at 0.5\%~points below. This can be partially explained by the frequent motion of the camera in the videos, making the optical flow noisy. Again, these results show that it is better to train a network end-to-end to combine two frames rather than feeding it optical flow directly. 

Overall, our RetinaNet-Double outperforms the baseline RetinaNet consistently on three datasets when trained in the same settings. This clearly shows that it is possible for a CNN to learn how to combine frames of a video, and that we should take advantage of that whenever possible. However, the lack of pre-trained weights makes it difficult to compare with current state-of-the-art benchmark results. We also demonstrated that using the optical flow directly does not help the network, as it might be too noisy and induce errors. 

\section{Discussion}
\label{resul}
The learning process can be sped up with transfer learning. Transfer learning was achieved on the KITTI dataset by first training a model on UA-Detrac, and fine-tuning on KITTI. Through transfer learning, the models can learn much faster than when training from scratch. On KITTI, it is about half the time it takes to train the model. 

\subsection{Detailed Analysis}
CNNs are often seen as black boxes with little knowledge on what they learn. It is therefore important to explain a few reasons why the proposed model achieves better performance than the baseline.

\begin{figure}[t]
\begin{center}
\includegraphics[width=0.6\linewidth]{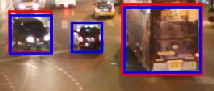}
\end{center}
   \caption{An example of RetinaNet-Double (blue) performing better on small objects than RetinaNet (red). }
\label{fig:small}
\end{figure}

\textbf{Small objects:} small objects are harder to locate precisely and classify than large objects for obvious reasons. There is a very large amount of small objects in the UA-Detrac dataset. Having two frames can help to address this challenge in multiple ways. If the object is moving, movement can be associated with the presence of an object. Combining features from two frames can help refine the features and thus classify more precisely. Figure~\ref{fig:small} shows a concrete example where RetinaNet-Double correctly detects a small car while the baseline does not. 

\textbf{Motion blur:} motion blur is frequent in traffic surveillance images. A blurry object can be harder to detect and classify. If the detector does not recognize it as one object of the pre-determined classes, the blurry object will be invisible to it. With two frames, movement can be associated with the presence of an object, and help overcome this challenge.  

\textbf{Occlusion:} occlusion happens quite often with traffic videos, either by other vehicles or by road furniture as can be seen in figure~\ref{fig:diff}. It is quite important to continue detecting vehicles even if they are partially hidden. Here, three elements can help us if we have access to more than one frame: 1) movement, 2) having more refined features and 3) the possibility that the vehicle is not hidden as much in the preceding frame. If the object is not as occluded in one of the two frames, then the network can learn to put more importance on the features of that frame.

\begin{figure}[t]
\begin{center}
\includegraphics[width=0.8\linewidth]{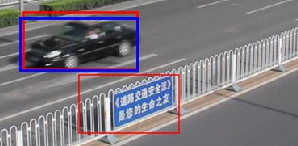}
\end{center}
   \caption{The baseline RetinaNet (red) wrongly classifies the billboard as a road user, while RetinaNet-Double (blue) does not.}
\label{fig:motion}
\end{figure}

\textbf{Motion:} Motion can serve to increase the probability of the presence of a road user, or in the opposite case the absence of motion can serve to decrease the probability of the presence of an object. In the example shown in figure~\ref{fig:motion}, the baseline model confused a billboard with a road user, while RetinaNet-Double did not. 

\subsection{Limitations of the Proposed Models}
The most obvious limitation of our models is that they rely on a pair of images to improve accuracy. In settings where every object is moving, this is beneficial. But to ensure that performance does not go down too much when objects stop moving, the models should be more thoroughly evaluated on other datasets. As long as the models see enough stationary objects during training, this should not be a problem. The key is to have a well-balanced training dataset. 

Another limitation is the need for pairs of images to run our model, which would become useless if only single frames are available. Nevertheless, in such situations, the model may simply be used with the same image twice as input. Preliminary tests in that regard show that performance does not go down significantly compared to the baseline when the model is used in that way. This could allow users to avoid re-training an entire other model just for the cases where pair of images are not available. 

\subsection{Generalization of the Proposed Models}
One of the distinct advantages of the proposed approach is that it generalizes to most object detection method. Since there is no change in the dimension of the features used by either the RPN of the two-stage approach or the sliding window of the one-stage approach, it could be used in almost any object detection method, and future experiments could determine for which it is the most useful. In fact, we could go further and provide pre-trained weights for pairs of images for multiple feature extractors to make experiments faster and more convenient. It would be very interesting for example to see a Faster R-CNN or R-FCN architecture use these models and compare with their baseline counterparts.

\section{Conclusion}
\label{conc}
Two novel models were introduced for road user detection and classification in videos, with improved performance on three datasets for RetinaNet-Double. These models were trained in different training settings, and compared with a baseline RetinaNet model. When trained from scratch, RetinaNet-Double achieves better mAP than the baseline, while RetinaNet-Flow shows that using optical flow for detection does not perform well and end-to-end learning of combining frames is better. A comparison is done with state of the art models on UAV, showing that when using pre-trained weights, our proposed architectures do not surpass the baseline model but compare well despite not using pre-trained weights for the first layer. Future work includes pre-training the proposed models on a large amount of video data, integrating the same ideas with other object detection models and using different backbone feature extractors.  

\section*{Acknowledgment}
We acknowledge the support of the Natural Sciences and Engineering Research Council of Canada (NSERC), [508883-17], and the support of Genetec. The authors would like to thank Paule Brodeur for insightful discussions.



%

\bibliographystyle{./IEEEtranBST/IEEEtran}
\bibliography{bib}

\end{document}